\documentclass{article}

\usepackage{microtype}
\usepackage{graphicx, amsmath}
\usepackage{booktabs} 
\usepackage{amsfonts} 
\usepackage{amssymb} 
\usepackage{multirow}
\usepackage{boldline}
\usepackage{array, makecell}
\usepackage{algorithm}
\usepackage{algorithmic}
\usepackage{hyperref}
\usepackage{url}
\usepackage{rotating}
\usepackage[utf8]{inputenc}

\usepackage{authblk}
\usepackage{comment}
\usepackage{numprint}
\usepackage[a4paper,includeheadfoot, vmargin=2cm]{geometry}
    
\providecommand{\keywords}[1]{\textbf{Keywords ---} #1}

\title{MultiVERSE: \\ a multiplex and multiplex-heterogeneous network embedding approach}

\author[1]{Léo Pio-Lopez}
\author[2]{Alberto Valdeolivas}
\author[1]{Laurent Tichit}
\author[1]{Élisabeth Remy}
\author[3,4]{Anaïs Baudot}

\affil[1]{Aix Marseille Univ, CNRS, Centrale Marseille, I2M, Marseille, France}
\affil[2]{Heidelberg University, Institute for Computational Biomedicine, Heidelberg, Germany}
\affil[3]{Aix Marseille Univ, INSERM, CNRS, MMG, Marseille, France}
\affil[4]{Barcelona Supercomputing Center, Barcelona, Spain}

\date{}

\begin{document}

\maketitle

\vskip 0.3in


\begin{abstract}
Network embedding approaches are gaining momentum to analyse a large variety of networks. Indeed, these approaches have demonstrated their efficiency through tasks such as community detection, node classification, and link prediction. However, very few network embedding methods have been specifically designed to handle multiplex networks, i.e. networks composed of different layers sharing the same set of nodes but having different types of edges. Moreover, to our knowledge, existing approaches cannot embed multiple nodes from multiplex-heterogeneous networks, i.e. networks composed of several multiplex networks containing both different types of nodes and edges.

In this study, we propose MultiVERSE, an extension of the VERSE framework using Random Walks with Restart on Multiplex (RWR-M) and Multiplex-Heterogeneous (RWR-MH) networks. MultiVERSE is a fast and scalable method to learn node embeddings from multiplex and multiplex-heterogeneous networks. 


We evaluate MultiVERSE on several biological and social networks and demonstrate its efficiency. MultiVERSE indeed outperforms most of the other methods in the tasks of link prediction and network reconstruction for multiplex network embedding, and is also efficient in link prediction for multiplex-heterogeneous network embedding. Finally, we apply MultiVERSE to study rare disease-gene associations using link prediction and clustering.\\

MultiVERSE is freely available on github at \url{https://github.com/Lpiol/MultiVERSE}.


\end{abstract}


\section{Introduction}
\label{intro}


Networks are powerful representations to describe, visualize, and analyse complex systems in many domains. Recently, machine learning techniques started to be used on networks, but 
these techniques have been developed for vector data and cannot be directly applied. A major challenge thus pertains to the encoding of high-dimensional graph-based data into a feature vector. Network embedding (also known as graph representation learning) provides a solution to this challenge and allows opening the complete machine learning toolbox for network analysis.


The high efficiency of network embedding approaches has been demonstrated in a wide range of applications such as community detection, node classification, or link prediction. Moreover, network embedding can leverage massive graphs, with millions of nodes \cite{hamilton2017representation}.
Thus, with the explosion of big data, network embeddings have been used to study many different networks, such as social \cite{liao2018attributed}, neuronal \cite{ma2017multi} and molecular networks \cite{nelson2019embed}.

So far, network embedding approaches have been mainly applied to monoplex networks (i.e. single networks composed of one type of nodes and edges) \cite{perozzi2014deepwalk, grover2016node2vec, hamilton2017representation}. 
Current technological advances however generate a large spectrum of data, which form large heterogeneous datasets. By design, monoplex networks are not suited to represent such diversity and complexity. Multi-layer networks, including multiplex \cite{Kivela2014} and multiplex-heterogeneous \cite{valdeolivas2018random} networks have been proposed to handle these more complex but richer heterogeneous interaction datasets.



\begin{figure}
\begin{center}
\includegraphics[width=1\linewidth]{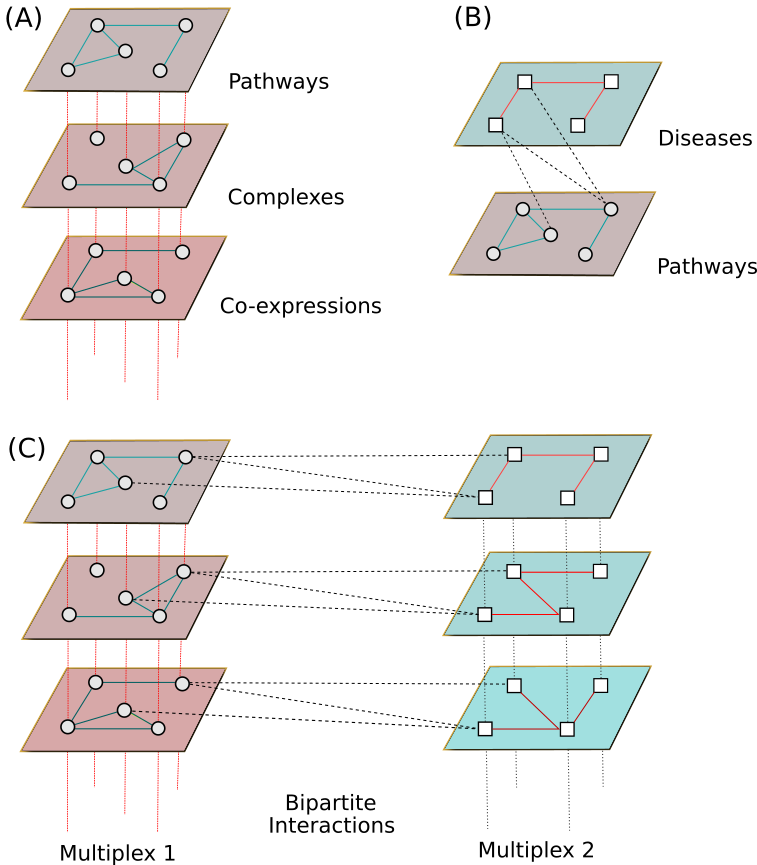} 
\end{center}
\caption{Illustrations of the different types of networks. (A) A multiplex network. The different layers share the same set of nodes but different types of edges. (B) A heterogeneous network. The two networks are composed of different types of nodes and edges, connected by bipartite interactions (black dashed lines). (C) A  multiplex-heterogeneous network composed of two multiplex networks. The multiplex networks are connected by bipartite interactions (dashed lines). For the sake of simplicity, the figure does not represent all the possible bipartite interactions (each layer of a given multiplex is in reality linked with every layer of the other multiplex).}
\label{net_All}
\end{figure}

Multiplex networks are composed of several layers, each layer being a monoplex network. All the layers share the same set of nodes, but their edges belong to different categories (Figure \ref{net_All}A). Multiplex representation is pertinent to depict the diversity of interactions between the same nodes. For instance, in a molecular multiplex network, the different layers could represent physical interactions between proteins, their belonging to the same molecular complexes or the correlation of expression of the genes across different tissues. Analogously, in social multiplex networks, a person can belong to different layers describing different types of relationships, such as friendships or common interests. 
 

A heterogeneous network is a multi-layer network in which each layer is a monoplex network with  its specific type of nodes and edges (Figure \ref{net_All}B). The two monoplex networks are connected by bipartite interactions, \textit{i.e.} edges linking the different types of nodes belonging to the two monoplex networks. Such heterogeneous networks have been studied in different research fields. For example, in network medicine, a  drug-protein target heterogeneous network has been constructed with a drug-drug similarity monoplex network, a protein-protein interaction monoplex network and bipartite interactions between drugs and their target proteins \cite{luo2017network}. In social science, citation networks are constructed with author-author and document-document monoplex networks connected by author-documents bipartite interactions, as in \cite{zhou2007co}.

A multiplex-heterogeneous network is a combination of heterogeneous and multiplex networks by connecting several multiplex networks through bipartite interactions (Figure \ref{net_All}C). 
The multiplex-heterogeneous structure is expected to provide a richer view on biological \cite{valdeolivas2018random}, social \cite{bagavathi2018multi} or other real-world systems describing complex relations among different components.  


Recently, different studies proposed  embedding approaches for multiplex networks \cite{zhang2018scalable, zitnik2017predicting, bagavathi2018multi, wilson2018fast} and heterogeneous networks \cite{dong2017metapath2vec, shi2018heterogeneous}. A recent method uses multiplex-heterogeneous information to embed one category of nodes \cite{dursun_gene_2020}. However, to our knowledge, no embedding methods are specifically dedicated to the embedding of nodes of different types from multiplex-heterogeneous networks.
In this paper, we present MultiVERSE, a fast, scalable and versatile embedding approach to learn node embeddings on multiplex and multiplex-heterogeneous networks. MultiVERSE is based on the VERSE framework \cite{tsitsulin2018verse}, and coupled with Random Walks with Restart on Multiplex (RWR-M) and on Multiplex-heterogeneous (RWR-MH) networks \cite{valdeolivas2018random}. 
Our contributions are the following:\begin{itemize}
\item   We propose an evaluation protocol in order to evaluate multiplex network embedding. It is based on 7 datasets in 4 disciplines (biological, neuronal, co-authorship and social networks), 6 embedding methods (and 4 additional link prediction heuristics), and two tasks: link prediction and a new protocol approach based on network reconstruction.
\item We demonstrate the higher performance of MultiVERSE over state-of-the-art network embedding methods in the tasks of link prediction and network reconstruction for multiplex network embedding.
\item We propose, to our knowledge, the first multiplex-heterogeneous network embedding  method (with an embedding of the different types of nodes).
\item We propose a method to evaluate multiplex-heterogeneous network embedding on link prediction. We demonstrate the efficiency of MultiVERSE on this task on two biological multiplex-heterogeneous networks.
\item We present a biological application of MultiVERSE for the study of gene-disease associations using link prediction and clustering.
\end{itemize}

\section{Related work in network embedding}

Network embedding relies on two key components: a similarity measure between pairs of nodes in the original network and a learning algorithm. Given a network and a similarity measure, the aim of network embedding is to learn a vector representation of the nodes in a lower dimension space, while preserving as much as possible the similarity. In the next sections we will present the state-of-the-art of monoplex, multiplex and multiplex-heterogeneous network embedding.

\subsection{Monoplex network embedding}

Many network embedding methods have been recently developed to study a large variety of networks, from biological to social ones. The classical method deepwalk \cite{perozzi2014deepwalk} inspired a series of methods such as node2vec \cite{grover2016node2vec} and LINE (for Large-scale Information Network Embedding) \cite{tang2015line}. Deepwalk uses truncated random walks to compute the node similarity in the network. Then, a combination of the skip-gram learning algorithm \cite{mikolov2013efficient} and hierarchical softmax \cite{mnih2009scalable} is used to learn the graph representations. Skip-gram is a model based on natural language processing. It intends to maximize the probability of co-occurrence of nodes within a walk, focusing on a window, i.e. a section of the path around the node. Node2vec \cite{grover2016node2vec} upgrades deepwalk by introducing negative sampling during the learning phase \cite{mikolov2013distributed}. Moreover, node2vec allows biasing the random walks towards depth or breadth-first random walks, in order to tune the exploration of the search space. LINE \cite{tang2015line} follows a different approach to optimize the embedding: it computes the node similarity using an adjacency-based proximity measure in association with negative sampling. 
Some embedding methods are based on matrix-factorization, such as GraRep \cite{cao2015grarep} or HOPE \cite{ou2016asymmetric}, and others on deep neural networks such as graph convolutional networks (GCN) \cite{kipf2016variational}.

These embedding methods have been applied to link prediction or node labelling tasks. Their performance rely upon multiple criteria such as the size of the network, its density, the embedding dimension or the evaluation metrics \cite{goyal2019benchmarks}. Overall, they have been designed to handle monoplex networks. However, we now have access to a richer representation of complex systems as multiplex networks, and some recent methods have explored the embedding of such multiplex networks.


\subsection{Multiplex network embedding}

The most straightforward approach to deal with multiplex networks is to merge the different layers into a monoplex network \cite{boccaletti2014structure}. However, this merging creates a new network with its own topology, and loses the topological features of the individual layers. This new topology is logically biased towards the initial topology of the denser layers \cite{de2016physics}. Different network embedding methods have been introduced in order to avoid merging multiplex network layers and take advantage of the multiplex structure \cite{zhang2018scalable, zitnik2017predicting, bagavathi2018multi, wilson2018fast}. Overall, these approaches are based on truncated random walks to compute the similarity in the multiplex network. Ohmnet \cite{zitnik2017predicting} relies on node2vec \cite{grover2016node2vec} and requires the definition of a hierarchy of layers to model dependencies between them. But usually, this layer hierarchy information is not known or easy to establish, particularly for multiplex networks such as social or molecular networks. The Scalable Multiplex Network Embedding (MNE) method \cite{zhang2018scalable} is also based on node2vec \cite{grover2016node2vec}. For each network node, it extracts one high-dimensional common embedding shared across all the layers of the multiplex network. In addition, MNE computes a lower-dimensional embedding for every node in each layer of the multiplex network. Multi-node2vec \cite{wilson2018fast} is another method based on node2vec that constructs the multiplex embedding with the random walks jumping from one layer to another. Multi-Net \cite{bagavathi2018multi} also proposes a random walks procedure in the multiplex network, inspired from \cite{guo2016levy}. Similarly to multi-node2vec, the random walks can jump from one layer to another. Multi-Net learns the embeddings using stochastic gradient descent. The performances of Ohmnet \cite{zitnik2017predicting}, Multi-net \cite{bagavathi2018multi} and MNE \cite{zhang2018scalable} have been compared in the context of network reconstruction \cite{bagavathi2018multi}. In this task, the aim is to reconstruct one layer of the multiplex network from the embeddings of the other layers. The results show better performances for Multi-net on a set of social and biological multiplex networks  \cite{bagavathi2018multi}.

\subsection{Multiplex-heterogeneous network embedding}

Some methods can perform the embedding of heterogeneous networks \cite{dong2017metapath2vec, shi2018heterogeneous}. 
A famous approach is metapath2vec \cite{dong2017metapath2vec}. It extends skip-gram to learn node embeddings for heterogeneous networks 
using meta-paths, which are predefined composite relations between different types of nodes. For instance, in the context of a drug-protein target heterogeneous network, the meta-path drug-protein target-drug in the network could bias the random walks to extract the information related to drug combinations. 

Nevertheless, to our knowledge, no approach is specifically dedicated to the embedding of different types of nodes from multiplex-heterogeneous networks. In the next section, we present formally MultiVERSE, a new method for multiplex and multiplex-heterogeneous network embedding relying on VERSE \cite{tsitsulin2018verse} and coupled with Random Walks with Restart extended to Multiplex (RWR-M) and Multiplex-Heterogeneous  graphs (RWR-MH) \cite{valdeolivas2018random}.

\section{MultiVERSE}
\label{sec:multiverse}

\begin{figure}[t]
\begin{center}
\includegraphics[width=1\linewidth]{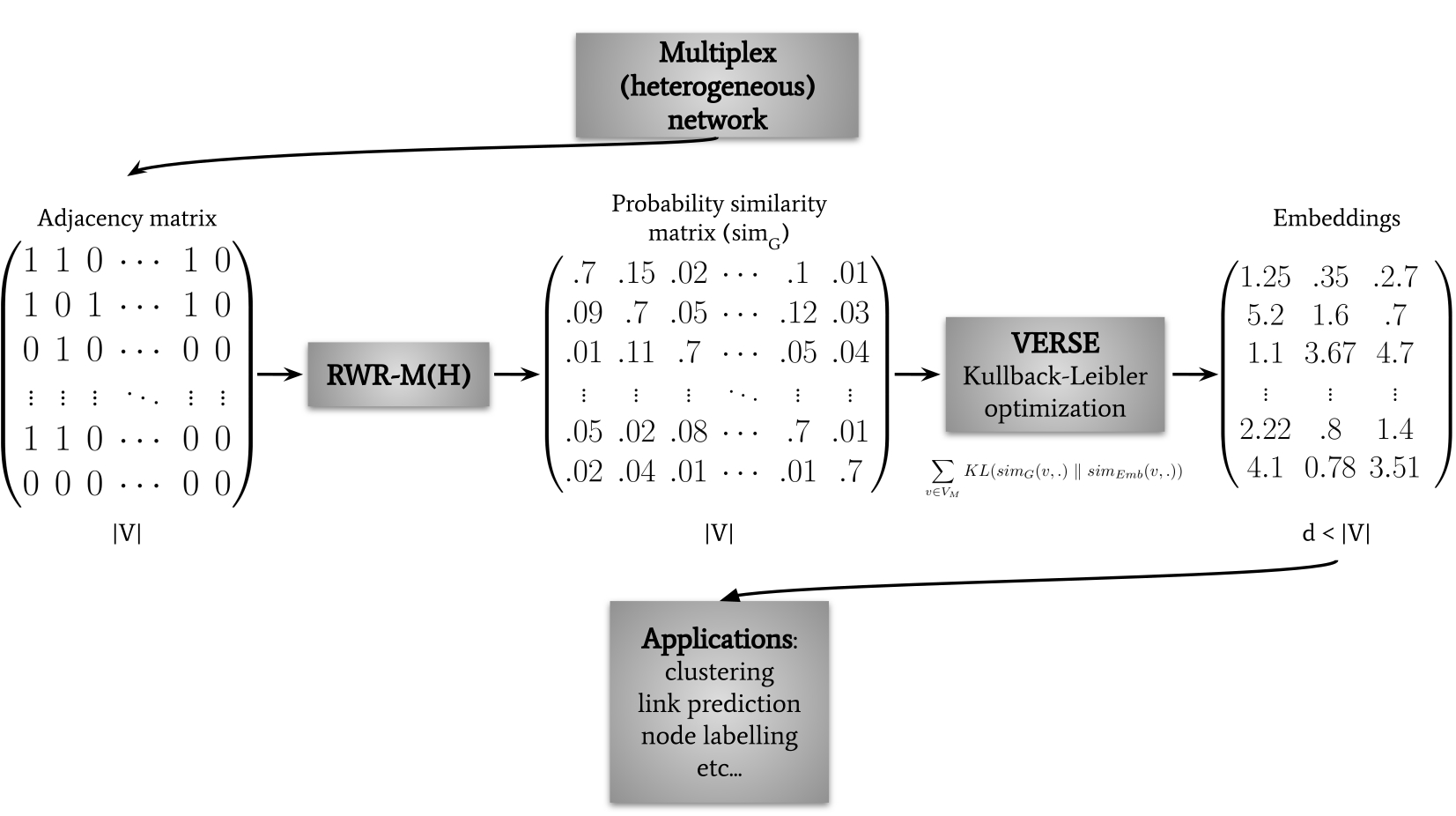} 
\end{center}
\caption{Overview of the MultiVERSE pipeline. Starting from a multiplex-heterogeneous network, we represent its structure through an adjacency matrix (size $|V| \times |V|$); we then compute a similarity matrix using Random Walk with Restart algorithm, and apply an optimized version of the VERSE algorithm to compute the embeddings. The resulting matrix of embeddings will be used for the applications.}
\label{general_pipeline_MultiVERSE}
\end{figure}

In this section, we present the key components of MultiVERSE: the VERSE general framework, the learning objective, and our particular implementation with Random Walk with Restart for Multiplex networks (RWR-M) and Random Walk with Restart for Multiplex-Heterogeneous networks (RWR-MH) (Figure \ref{general_pipeline_MultiVERSE}). We finally describe the MultiVERSE algorithm.

\subsection{VERSE: a general framework for network embedding}


The aim of VERSE network embedding is to learn a low-dimensional nonlinear representation $w_i$ of the nodes $v_i$ to a $d$-dimensional continuous vector, where $d < n$, using Kullback-Leibler optimization \cite{tsitsulin2018verse}. We denote $d$ the dimension of the embedding space, and $n$ the dimension of the adjacency matrix. VERSE was originally developed for the embedding of monoplex networks \cite{tsitsulin2018verse}. The VERSE framework is nevertheless general and versatile enough to be expanded to multiplex and multiplex-heterogeneous networks.

\subsubsection{Similarity distributions}

Consider an undirected graph $G=(V, E)$ with $V=\{v_i, i=1, \dots,n\}$ the set of nodes ($|V| = n$), and $E \subseteq V \times V$ the set of edges, and $sim_G\,:\,V \times V \rightarrow \mathbb{R}$  a given similarity measure on $G$ such that
\begin{equation}
\forall v \in V, \ \sum_{u\in V} sim_{G} (v, u) = 1\,.
\end{equation}
Hence, the similarity for any node $v$ is expressed as a probability distribution $sim_G(v,.)$.
 We note $w_i$ the vector representation of node $i$ in the embedding space ($W$ is a $(n \times d)$-matrix). 

The (non-normalized) distance or similarity between two nodes embeddings $w_u$ and $w_v$ is defined as the dot product $w_u \cdot w_v^T$. Using the softmax function, we obtain the normalized similarity distribution in the embedding or vector space:
\begin{equation}
sim_{Emb}(v, .) = \frac{exp(w_v \cdot w^T)}{\sum_{i=1}^n exp (w_v \cdot w_i)}\,.
\label{simE}
\end{equation}
 
Finally, the output of any network embedding method is a matrix of embeddings $W$ such as, $\forall v \in V$,  $sim_{Emb}(v, .) \approx sim_G(v, .)$. This requires a learning phase, which is described in the next section.

\subsubsection{Learning objective}

This step updates the embeddings at each iteration in order to project $sim_G$ into the embedding space leading to the preservation of the topological structure of the graph. In the framework of VERSE, as $sim_{Emb}$ and $sim_G$ are both probability distributions, this optimization phase aims to minimize the Kullback-Leibler divergence (KL-divergence) between these two similarities:

\begin{equation}
\sum_{v \in V_M} KL(sim_G(v, .)~\Vert~ sim_{Emb}(v,.))
\label{KL}
\end{equation}

We can keep only the parts related to $sim_{Emb}$ as it is the target to optimize and $sim_G$ is constant. This leads to the following objective function:

\begin{equation}
\mathcal{L} = - \sum_{v \in V_M} sim_G(v, .)~log(sim_{Emb}(v,.))
\label{KL_log}
\end{equation}

$sim_{Emb}$ is defined as a softmax function and needs to be normalized over all the nodes of the graph at each iteration, which is computationally heavy. Therefore, following the VERSE algorithm \cite{tsitsulin2018verse}, we used Noise Contrastive Estimation (NCE) to compute this objective function \cite{gutmann2010noise, mnih2012fast}. NCE trains a binary classifier to distinguish node samples coming from the distribution of similarity in the graph $sim_G$ and those generated by a noise distribution $Q$. We define $D$ as the random variable representing the classes, $D=0$ for a node if it has been drawn from the noise distribution $Q$ or $D=1$ if it has been drawn from the empirical distribution and $\mathbb{E}$ is the expected value. With $u$ a node drawn from $\mathcal{P}$ and $v$ drawn from $sim_G (u, .)$, with NCE we draw $s < n$ negative samples $v_{neg}$ from $Q(u)$.

In this framework, the objective function becomes the negative log-likelihood that we want to minimize via logistic regression:


\begin{equation}
\begin{split}
\mathcal{L}_{NCE} &= \sum_{\substack{u \sim \mathcal{P} \\ v \sim sim_G(u,.) }} \Big[ log P_W (D=1 ~|~ sim_{Emb}(u,v)) \\
& + s.\mathbb{E}_{v_{neg} \sim Q(u)} log P_W(D=0 ~|~	sim_{Emb}(u, \widetilde{v}))  \Big]
\end{split}
\label{NCE}
\end{equation}

where $P_W$ is computed as the sigmoid ($\sigma (x) = (1 + e^{-x})^{-1}$) of the dot product of the embeddings $w_u$ and $w_v$, and $sim_{Emb}(u,.)$ is computed without normalization. It has been proven that the derivative of NCE converges to gradient of cross-entropy when $s$ increases, but in practice small values work well \cite{mnih2012fast}. Therefore, we are minimizing the KL-divergence from $sim_G$.

Overall, VERSE is a general framework for network embedding with the only constraint that $sim_G$ must be defined as a probability distribution. In this work, we computed $sim_G$ using Random Walks with Restart on Multiplex (RWR-M) and Random Walks with Restart on Multiplex-Heterogeneous (RWR-MH) networks \cite{valdeolivas2018random}. We describe this particular implementation in the next section.


\subsection{Random Walk with Restart on Multiplex and Multiplex-Heterogeneous networks}

\subsubsection{Random Walk (RW) and Random Walk with Restart (RWR)}

Let us consider a finite graph, $G = (V,E)$, with adjacency matrix $A$. In a classical RW, an imaginary particle starts from a given initial node, $v_{0}$. Then, the particle moves to a randomly selected neighbour of $v_{0}$ with a probability defined by its degree. We can define $p_{t}(v)$ as the probability for the random walk to be at node $v$ at time $t$. Therefore, the evolution of the probability distribution, $\textbf{p}_{t}=(p_{t}(v))_{v \in V}$, can be described as follows:
\begin{equation} 
\label{eq:01}
\textbf{p}^{T}_{t+1}=M\textbf{p}^{T}_{t}\,
\end{equation}
\noindent where $M$ denotes a transition matrix that is the column normalization of $A$. The stationary distribution of Equation~(\ref{eq:01}) represents the probability for the particle to be located at a specific node for an infinite amount of time \cite{Lovasz1993}.

Random Walk with Restart (RWR) additionally allows the particle to jump back to the initial node(s), known as seed(s), with a probability $r \in (0,1)$ at each step. In this case, the stationary distribution can be interpreted as a measure of the proximity between the seed(s) and all the other nodes in the graph. We can formally define RWR by including the restart probability in Equation~(\ref{eq:01}):

\begin{equation} 
\label{eq:02}
\textbf{p}^{T}_{t+1}=(1-r)M\textbf{p}^{T}_{t} + r\textbf{p}^{T}_{0}
\end{equation}
The vector $\textbf{p}_{0}$ is the initial probability distribution. Therefore, in $\textbf{p}_{0}$, only the seed(s) have values different from zero. Equation~(\ref{eq:02}) can be solved in a iterative way \cite{valdeolivas2018random}. 

In our previous work, we expanded the Random Walk with Restart algorithm to Multiplex (RWR-M) and Multiplex-Heterogeneous networks (RWR-MH) \cite{valdeolivas2018random}. Below, we show how the output of RWR-M and RWR-MH can easily be adapted to produce $sim_G$, the required input for the VERSE framework. 

\subsubsection{Random Walk with Restart on Multiplex networks (RWR-M)}

We define a multiplex graph as a set of $L$ undirected graphs, termed layers, which share the same set of $n$ nodes \cite{Kivela2014,DeDomenico2014}. The different layers, $\alpha=1, \dots ,L$, are defined by their respective $n \times n$ adjacency matrices, 
$A^{[\alpha]}=({A^{[\alpha]}(i,j)})_{i,j=1,\dots,n}$. $A^{[\alpha]}(i,j) =1$ if node i and node j are connected on layer $\alpha$, and 0 otherwise \cite{Battiston2014}. We do not take into account potential self-interactions and therefore set ${A^{[\alpha]}(i,i)} = 0 \ \forall \ i=1,\dots,n$. In addition, we consider that $v_i^{\alpha}$ represents the node $i$ in layer $\alpha$.
 
Thus, we can represent a multiplex graph by its adjacency matrix:
\begin{equation}
\textbf{A }= {A^{[1]}, \dots ,A^{[L]}}\,
\end{equation}
\noindent and define it as $G_{M}=(V_{M},E_{M})$, where:
\begin{gather*}
V_{M} = \left \{ v_{i}^{\alpha}, \ i=1,\dots,n , \ \alpha=1,\dots,L  \right \}\,, \\
\begin{split}
E_{M} = \left \{(v_{i}^{\alpha},v_{j}^{\alpha}), \ i,j=1,\dots,n , \ \alpha=1,\dots,L, \ {A^{[\alpha]}(i,j)} \neq 0 \right \} \bigcup \\
	\left \{(v_{i}^{\alpha},v_{i}^{\beta}), \ i=1,\dots,n, \ \alpha \neq \beta \right \}\, . \\ 
\end{split}
\end{gather*}

RWR-M should ideally explore in parallel all the layers of a multiplex graph to capture as much topological information as possible. Therefore, a particle located in a given node, $v_{i}^{\alpha}$, may be able to either walk to any of its neighbours within the layer $\alpha$ or to jump to its counterpart node in another layer, $v_{i}^{\beta}$ with $\beta \neq \alpha$ \cite{DeDomenico2013}. Additionally, the particle can restart in the seed node(s) on any layer of the multiplex graph. In order to match these requirements, we previously defined a multiplex transition matrix and expanded the restart probability vector, allowing us to apply Equation~(\ref{eq:01}) on multiplex graphs \cite{valdeolivas2018random}.

In this study, we independently run the RWR-M algorithm $n$ times, using each time a different node as seed. This allows measuring the distance from every single node to all the other nodes in the multiplex graph. This node-to-node distance matrix is actually a probability distribution describing the particle position in the steady state, where $\sum_{u\in V_{M}} sim_{G} (v, u) = 1 \; \forall \: v \in V_{M}$, therefore fulfilling the requirements of the VERSE input. We set the RWR-M parameters to the same values used in our original study ($r = 0.7$, $\boldsymbol{\tau} = (1/L,1/L,\dots,1/L)$, $\delta= 0.5$) \cite{valdeolivas2018random}.

\subsubsection{Random Walk with Restart on Multiplex-Heterogeneous networks (RWR-MH)}

A heterogeneous graph is composed of two graphs with different types of nodes and edges. In addition, it also contains a bipartite graph in order to link the nodes of different type (bipartite edges) \cite{Lee2013}.
In our previous study \cite{valdeolivas2018random}, we described how to extend the RWR to a graph which is both multiplex and heterogeneous. However, this study considered only one multiplex graph in the multiplex-heterogeneous graph. For the present work, we additionally expanded RWR-MH to a complete multiplex-heterogeneous graph, i.e. both components of the heterogeneous graph can be multiplex (Figure \ref{net_All}, C), based on the work of \cite{Dursun2019CompleteRWRMH}. 
Let us consider a $L$-layers multiplex graph, $G_{M}=(V_{M},E_{M})$, with $n \times L$ nodes, $V_{M} = \left \{v_{i}^{\alpha}, \ i=1,\dots,n , \ \alpha=1,\dots,L \right \}$. We also define a second $L$-layers multiplex graph, with $m \times L$ nodes, $U_{M} = \left \{u_{j}^{\alpha}, \ j=1,\dots,m , \ \alpha=1,\dots,L \right \}$. We additionally need a bipartite graph $G_{B}=(V_{M} \cup U_{M}, E_{B})$ with $E_{B} \subseteq V_{M} \times U_{M}$. The edges of the bipartite graph only connect pairs of nodes from the different sets of nodes, $V_{M}$ and $U_{M}$. It is to note that the bipartite edges should link nodes with every layer of the multiplex graphs. We therefore need $L$ identical bipartite graphs, $G_{B}^{[\alpha]}= (V_{M} \cup U_{M}, E^{[\alpha]}_{B})$ to define the multiplex-heterogeneous graph. We can then describe a multiplex-heterogeneous graph, $G_{MH}=(V_{MH}, E_{MH})$, as:
\begin{gather*}
V_{MH} = \left \{ V_{M} \cup U_{M} \right \} \\
E_{MH} = \left \{\cup_{\alpha=1,\dots,L}E^{[\alpha]}_{B} \cup E_{V_{M}} \cup E_{U_{M}} \right \} 
\end{gather*}

In the RWR-MH algorithm, the particle should be allowed to move in any of the multiplex graphs as described in the RWR-M section. In addition, it may be able to jump from a node in one multiplex graph to the other multiplex graph following a bipartite edge. We also have to bear in mind that the particle could now restart in different types of node(s), i.e. we can have seed(s) of different category (see Figure \ref{net_All}, C). We accordingly defined a multiplex-heterogeneous transition matrix and expanded the restart probability vector. This gave us the opportunity to extent and apply Equation~(\ref{eq:01}) on multiplex-heterogeneous graphs \cite{valdeolivas2018random, Dursun2019CompleteRWRMH}.

In the context of MultiVERSE, we independently run the RWR-MH algorithm $n+m$ times. In each execution, we select a different seed node until all the nodes from both multiplex graphs have been used as individual seeds. As a result, we can define a node-to-node distance matrix matching VERSE input criteria, i.e $\sum_{u\in V_{MH}} sim_{G} (v, u) = 1 \; \forall \: v \in V_{EM}$. We set the RWR-MH parameters to the same values used in the original study ($r = 0.7$, $\boldsymbol{\tau} = (1/L,1/L,\dots,1/L)$, $\delta= 0.5$, $\lambda=0.5$, $\eta=0.5$) \cite{valdeolivas2018random}.

\subsection{MultiVERSE algorithm}


Algorithm \ref{multiverse_algo} presents the pseudo-code of MultiVERSE based on RWR on multiplex and multiplex-heterogeneous networks \cite{valdeolivas2018random} and Kullback-Leibler optimization from the VERSE algorithm \cite{tsitsulin2018verse}.

Our implementation of VERSE with NCE is slightly different from the original. We perform first the RWR-M or RWR-MH for all the nodes of the network in order to obtain the similarity distribution $sim_{G_M}$. The output of this step is the probability matrix $\overline{\textbf{p}}$, where $\overline{\textbf{p}}_u$ is the probability vector representing the similarities between $u$ and all the other nodes. The matrix of the embedded representation of the nodes, $W$, is randomly initialized. For each iteration, from one node $u$ sampled randomly from a uniform distribution $\mathcal{U}$, we filter the probability vector $\overline{\textbf{p}}_u$. We keep the $N_{max}$ highest probabilities because the shape of the distribution of probabilities falls very fast to very low probabilities.  Doing so, we can speed up the calculation by filtering out the lowest probabilities and by reducing the size of the similarity matrix.  We normalize this resulting probability vector $\dot{\textbf{p}}_u$, and sample one node $v$ according to its probability in $\overline{\textbf{p}}_u$. We set empirically the parameter $N_{max} = 300$ for large networks (number of nodes exceeding $5000$). For smaller networks, we set this parameter to $10\%-20\%$ of the number of nodes of the network, depending on the shape of the distribution. These two steps (lines $6$ and $7$) were not in the original VERSE.
We parallelized the repeat loop (line 4) and added a parallelized for loop after line 5 in order to run the code from line 6 to 12 in parallel $P$ times. In our simulations, we set $P=100$.



\begin{algorithm}
\begin{algorithmic}[1] 
\STATE {\textbf{Input:} a multiplex graph, $N_{max}$, $s$ } 
\STATE {$W \gets \mathcal{N}(0,1)$} 
\STATE {$\overline{\textbf{p}} \gets \text{RWR-M(H)}(G_M)$ } 
\REPEAT
\STATE {$u \sim \mathcal{U}$} 
\STATE {$\dot{\textbf{p}}_u = \text{Normalize}(\overline{\textbf{p}}_u(1,...,N_{max}))$}
\STATE {$v_{pos} \sim \dot{\textbf{p}}_u $} 
\STATE {$W_u, W_{v_{pos}} \gets Update (u,~ v_{pos} ,~ 1,~ bias_{pos})$}
	\FOR{i=1,~...~,~s} 
		\STATE {$v_{neg} \sim \mathcal{Q}(u)$} 
		\STATE {$W_u, W_{v_{neg}} \gets Update (u, ~v_{neg} , ~0,~ bias_{neg})$}
	\ENDFOR 
\UNTIL{Maximum step reached} 
\end{algorithmic}
\caption{MultiVERSE algorithm}
\label{multiverse_algo}
\end{algorithm}

Then, we update $W_u$ and $W_v$ according to algorithm \ref{update_algo} by reducing their distances in the embedding space. We added the bias for NCE: $bias_{pos}=log(N)$ and $bias_{neg}=log(N/s)$.

\begin{algorithm}
\begin{algorithmic}[1] 
\STATE {\textbf{Input:} $u$, $v$, $D$,~$bias$,~ $lr$ } 
\STATE {$g \gets [D-\sigma(W_u \cdot W_v - bias)]* lr$}
\STATE {$W_u \gets g\cdot W_v$}
\end{algorithmic}
  \caption{Update}
  \label{update_algo}
\end{algorithm}

Then, $s$ negative nodes are sampled from $\mathcal{Q}(u)$ and we update the corresponding embeddings by increasing their distances in the embedding space. The update can also be seen as the training part with $lr$ as the learning rate of the binary classifier of the NCE estimation as described in equation \ref{NCE}. The whole process is repeated until the maximum steps are reached. 

MultiVERSE is freely available on github at \url{https://github.com/Lpiol/MultiVERSE}.

\section{Evaluation protocol}
 \label{sec:eval}
 
We propose a benchmark to compare the performance of MultiVERSE and other embedding methods for multiplex and multiplex-heterogeneous networks. The performances are evaluated through link prediction for both multiplex and multiplex-heterogeneous networks, and with network reconstruction for multiplex networks.
 

\subsection{Evaluation of multiplex network embedding}

In the next sections, we describe the datasets, the evaluation tasks and the methods used for evaluations.

\subsubsection{Multiplex network datasets}
\label{ssec:ref_H}

We used 7 multiplex networks (2 molecular, 1 disease, 1 neuronal, 1 co-authorship and 2 social networks) to evaluate the different approaches of multiplex network embedding. The networks CKM, LAZEGA, C.ELE, ARXIV, and HOMO have been extracted from the CoMuNe lab database \url{https://comunelab.fbk.eu/data.php}. We constructed the other two networks, DIS and MOL. A description of each of these multiplex networks follows. The number of nodes and edges of the different layers are detailed in Table \ref{sample-table}.

\begin{itemize}
\itemsep0em 

\item \textbf{CKM physician innovation (CKM)} is a multiplex network describing how physicians in four towns in Illinois used the new drug tetracycline \cite{coleman1957diffusion}. It is composed of 3 layers corresponding to three questions asked to the physicians: i) to whom do you usually turn when you need information or advice about questions of therapy? ii) who are the three or four physicians with whom you most often find yourself discussing cases or therapy in the course of an ordinary week -- last week for instance? iii) would you tell me the first names of your three friends whom you see most often socially?

\item \textbf{Lazega network (LAZEGA)} is a multiplex social network composed of 3 layers based on co-working, friendship and advice between partners and associates of a corporate law partnership \cite{emmanuel2001collegial}. 

\item \textbf{Caenorabidis Elegans connectome (C.ELE)} is a neuronal multiplex network composed of 3 layers corresponding to different synaptic junctions \cite{chen2006wiring, de2015muxviz}: electrical, chemical poladic and chemical monadic.

\item \textbf{ArXiv network (ARXIV)} is composed of 8 layers corresponding to different ArXiv categories. The dataset has been restricted to papers with 'networks' in the title or abstract, up to May 2014 \cite{de2015identifying}. The original data from the CoMuNe Lab database is divided in 13 layers. We extracted the 8 layers (1-2-3-5-6-8-11-12) containing more than 1000 edges.

\item \textbf{Homo sapiens network (HOMO)} is composed of 4 layers extracted from the original network on CoMuNe Lab \cite{de2015muxviz}, keeping physical association, direct interaction, association and co-localization layers. The data are initially extracted from BioGRID \cite{stark2006biogrid}

\item \textbf{Disease multiplex network (DIS)} has been constructed, composed of 3 layers: i) A disease-disease network based on a projection of a disease-drug network from the Comparative Toxicogenomics Database (CTD) \cite{davis2019comparative} extracted from BioSNAP \cite{biosnapnets}. In this network, an edge between two diseases is created if the Jaccard Index between the neighborhoods of the two nodes in the original bipartite network is superior to $0.4$.  Two diseases are thereby linked if they share a similar set of drugs. This projection has been done using NetworkX \cite{hagberg2008exploring}. ii) A disease-disease network where the edges are based on shared symptoms. The network has been constructed from the bipartite disease-symptoms network from \cite{zhou2014human}. Similarly to \cite{zhou2014human}, we use the cosine distance to compute the symptom-based diseases similarity for this network. We kept for the disease-disease network all interactions with a cosine distance superior to $ 0.5$ iii) A comorbidity network from epidemiological data extracted from \cite{jensen2014temporal}.

\item \textbf{Human molecular multiplex network (MOL)} is a  molecular network, consisting of 3 layers: i) A protein-protein interaction (PPI) layer corresponding to the fusion of 3 datasets: APID (\url{apid.dep.usal.es}) (Level 2, human only), Hi-Union and Lit-BM (\url{http://www.interactome-atlas.org/download}). ii) A pathways layer extracted from NDEx \cite{pratt2015ndex} and corresponding to the human Reactome data \cite{croft2014reactome}. iii) A molecular complexes layer constructed from the fusion of Hu.map \cite{drew2017integration} and Corum \cite{giurgiu2019corum}, using OmniPathR \cite{turei2020omnipath}.

\end{itemize}


\begin{table}[t]
\begin{center}
\begin{small}
\begin{tabular}{c c c c}
\thead{Dataset}  & \thead{Layers} & \thead{Nodes} & \thead{Edges} \\ 
\midrule
\multirow{3}{*}{CKM}   		                     & 1 & 215 & 449 \\
					 					 		 & 2 & 231 & 498 \\ 
					 					 		 & 3 & 228 & 423 \\ \hline
\multirow{3}{*}{LAZEGA}                          & 1  & 71 & 717 \\ 
						 						 & 2 & 69 & 399 \\ 
						 						 & 3 & 71 & 726 \\ \hline
\multirow{3}{*}{C.ELE}                           & 1 & 253 & 514 \\ 
						 						 & 2 & 260 & 888 \\ 
						 						 & 3 & 278 & 1703 \\ \hline
\multirow{8}{*}{ARXIV}                           & 1 & 1558 & 3013 \\ 
 												 & 2 & 5058 & 14387 \\ 
 												 & 3 & 2826 & 6074 \\ 
											 	 & 4 & 1572 & 4423 \\ 
 												 & 5 & 3328 & 7308 \\
 												 & 6 & 1866 & 4420 \\ 
 												 & 7 & 1246 & 1947 \\ 
 												 & 8 & 4614 & 11517 \\ \hline
\multirow{4}{*}{HOMO}                            & 1 & 12345 & 48528 \\
											     & 2 & 14770 & 83414 \\ 
											     & 3 & 1626 & 1953 \\ 
										         & 4 & 5680 & 18381 \\ \hline
\multirow{3}{*}{DIS}                             & 1 & 3891 & 117527 \\ 
											     & 2 & 4155 & 101104 \\ 
											     & 3 & 434 & 3137 \\ \hline
\multirow{3}{*}{MOL}                             & 1 & 14704 & 122211 \\ 
												 & 2 & 7926 & 194500 \\ 
												 & 3 & 8537 & 63561 \\ 
\end{tabular}
\end{small}
\end{center}
\caption{Description of the 7 multiplex networks used for the evaluation protocol. }
\label{sample-table}
\end{table}

\subsubsection{Methods implemented for comparisons}

We compare MultiVERSE with 6 methods designed for monoplex network embedding (deepwalk, node2vec, LINE) and multiplex network embedding (Ohmnet, MNE, Multi-node2vec), and 4 link prediction heuristic scores (only in the link prediction task).

\subsubsection*{Monoplex network embedding methods}
\begin{itemize}

\item \textbf{deepwalk} \cite{perozzi2014deepwalk}: This method is based on non-biased random walks, and apply the skip-gram algorithm \cite{mikolov2013efficient} to learn the embeddings. We set the context window to 10, and the number of random walks to start at each node to 10.

\item \textbf{node2vec} \cite{grover2016node2vec}: This method is an extension of deepwalk with a pair of parameters $p$ and $q$ that biases the random walks for Breadth-first Sampling or Depth-first Sampling. We set $p=2$ and $q = 1$ to promote moderate explorations of the random walks from a node, as stated in \cite{grover2016node2vec}. We set the other parameters as for deepwalk.

\item \textbf{LINE} \cite{tang2015line}: LINE is not based on random walks, but computes the similarities using an adjacency-based proximity measure in association with negative sampling. It approximates the first and second order proximities in the network from one node. First order proximity refers to the local pairwise proximity between the nodes in the network (only neighbours), and second order proximity look for nodes sharing many connections. We set the negative ratio to 5.

\end{itemize}

\subsubsection*{Multiplex network embedding methods}

\begin{itemize}

\item \textbf{OhmNet} \cite{zitnik2017predicting}: This approach takes into account the multi-layer structure of multiplex networks. It is a random walk-based method that uses node2vec to learn the embeddings layer by layer. We applied the same parameters as in node2vec. The user has to define a hierarchy between layers. We created a 2-level hierarchy for all multiplex networks with first layer as the higher in the hierarchy and the other layers are defined at the second level of the hierarchy, in the same way as \cite{bagavathi2018multi}.

\item \textbf{MNE} \cite{zhang2018scalable}: This method is also designed for multiplex networks and uses node2vec to learn the embeddings layer by layer. For each node, MNE computes a high-dimensional common embedding and a lower-dimensional additional embedding for each type of relation of the multiplex network. The final embedding is computed using a weighted sum of these two high-dimensional and low-dimensional embeddings. We used the default parameters (\url{https://github.com/HKUST-KnowComp/MNE}).

\item \textbf{Multi-node2vec} \cite{wilson2018fast}: This multiplex network embedding method is also based on node2vec. The random walks can jump to different layers and explore in this way the multiplex neighborhood. The length of the random walks is set to $100$.

\end{itemize}

We used OpenNE (\url{https://github.com/thunlp/OpenNE}) to implement deepwalk, node2vec and LINE. The other methods have been implemented from the source code associated to the different publications.

\subsubsection*{Link prediction heuristics}

In order to evaluate the relevance of the aforementioned network embedding methods, we also compared them with four classical and straightforward link prediction heuristic scores for node pairs \cite{grover2016node2vec}. Table \ref{graph_operator} provides formal definitions of these heuristic scores.

\begin{table}[t]
\begin{center}
\begin{small}
\begin{tabular}{lc}
Score & Definition \\
\midrule
Jaccard Coefficient (JC)   & $\frac{|\mathcal{N}(u) \cap \mathcal{N}(v)|}{|\mathcal{N}(u) \cup \mathcal{N}(v)|}$  \\[3mm]
Common neighbours (CN)  & $|\mathcal{N}(u) \cap \mathcal{N}(v)|$ \\[3mm]
Adamic Adar (AA)   & $\sum_{t \in |\mathcal{N}(u) \cap \mathcal{N}(v)|}\frac{1}{\log |\mathcal{N}(t)|} $\\[3mm]
Preferential attachment (PA)   & $|\mathcal{N}(u)| . |\mathcal{N}(v)|$ \\
\end{tabular}
\end{small}
\end{center}
\caption{Definition of the heuristic scores of a link $(u,v)$  in the graph $G(V,E)$. $\mathcal{N}(u)$ denotes the set of neighbour nodes of node $u \in V$ in $G(V,E)$.}
\label{graph_operator}
\end{table}

\subsubsection{Evaluation tasks}
\label{sssec:eval}

On multiplex networks, we evaluate the different methods by measuring their performances in two different tasks: link prediction and network reconstruction. For all the evaluations, we set the embedding dimension to $d=128$ as in \cite{grover2016node2vec, zitnik2017predicting, perozzi2014deepwalk}, and used the package EvalNE v0.3.1 \cite{mara2019evalne}.

\subsubsection*{From node embeddings to edges}

MultiVERSE and the other embedding methods allow learning vector representations of nodes from networks. We aim here to test their performance on link prediction and network reconstruction. We hence need to predict whether an edge exists between every pairs of node embeddings. To do so, given two nodes $u$ and $v$, we define an operator $\circ$ over the corresponding embeddings $f(u)$ and $f(v)$. This gives $g: V \times V \rightarrow \mathbb{R}^d$, with $d$ the dimension of the embeddings, $V$ the set of nodes and $g(u,v)=f(u) \circ f(v)$. Our test network contains both true and false edges (present and absent edges, respectively). We apply five different operators $\circ$: Hadamard, Average, Weighted-L1, Weighted-L2 and Cosine (Table \ref{binary_operators})).

The outputs of the embedding operators are used to feed a binary classifier for the evaluation tasks. This classifier aims to predict if there is an edge or not between two nodes embeddings.
Similarly, we use the output of the four link prediction heuristic scores  described in Table \ref{graph_operator} with a binary classifier to predict edges in a multiplex network.

\begin{table}[t]
\begin{center}
\begin{small}
\begin{tabular}{lccr}
Operators & Symbol & Definition \\
\midrule
Hadamard     & $\boxdot$   & $[f(u) \boxdot f(v)]_i = \frac{f_i(u) * f_i(v)}{2}$\\[2mm]
Average   & $\boxplus$   & $[f(u) \boxplus f(v)]_i = f_i(u) + f_i(v)$ \\[2mm]
Weighted-L1   & $\parallel . \parallel_{\overline{1}}$ & $\parallel f_i(u) . f_i(v) \parallel_{\overline{1}}i = \mid f_i(u) - f_i(v) \mid$ \\[2mm]
Weighted-L2   & $\parallel . \parallel_{\overline{2}}$ & $\parallel f_i(u) . f_i(v) \parallel_{\overline{2}}i = \mid f_i(u) - f_i(v) \mid^2$ \\[2mm]
Cosine     &  $cos$   & $cos[f(u), f(v)]_i = \frac{f_i(u) * f_i(v)}{\parallel f_i(u)\parallel \parallel f_i(v) \parallel}$\\
\end{tabular}
\end{small}
\end{center}
\caption{Embedding operators used to predict edges in the tasks of link prediction and network reconstruction. The definitions describe the $i^{th}$ components of $g(u,v)$.}
\label{binary_operators}
\end{table}

\subsubsection*{Link prediction}

\begin{figure}[t]
\begin{center}
\includegraphics[width=1\linewidth]{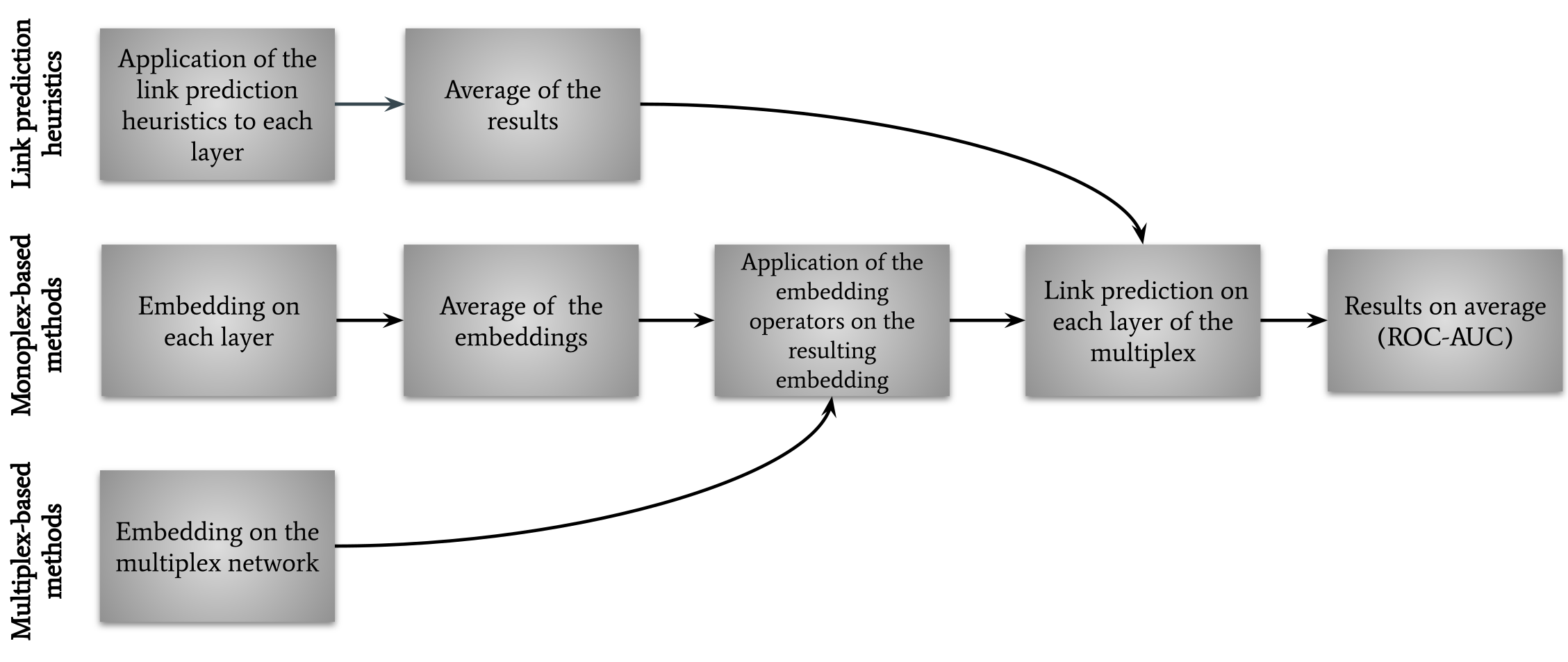} 
\end{center}
\caption{General approach for link prediction on multiplex networks: (top) for the link prediction heuristics, we apply them to each layer and average them across all layers~; (center) for monoplex-based methods, we embed each layer with the given method, then average it~; (bottom) for multiplex-based methods, we apply the specific embedding method to the network. The embedding operators are then applied to monoplex- and multiplex-based method embeddings. The three types of methods are finally evaluated for link prediction using a binary classifier and a ROC-AUC is computed.}
\label{pipeline_linkpred}
\end{figure}

We first evaluate the performance of the different methods to predict links removed from the original multiplex networks (Figure \ref{pipeline_linkpred}).
We remove 30\% of the links in each layer of the original networks. We applied the Andrei Broder algorithm  \cite{broder1989generating} in order to randomly select the links to be removed while keeping a connected graph in each layer. This step provides the multiplex training network, to which we apply the 3 categories of methods (see Figure \ref{pipeline_linkpred}):

\begin{itemize}
\item The methods specifically designed for monoplex network embedding (node2vec, deepwalk and LINE) are applied individually on each layer of the multiplex networks. We thereby obtain one embedding per layer and average them (arithmetic mean) in order to obtain a single embedding for each node. We then apply the embedding operators. We refer to these approaches in the results section as node2vec-av, deepwalk-av and LINE-av. 

\item Methods specifically designed for multiplex network embedding (Ohmnet, MNE, Multi-node2vec) are applied directly on the training multiplex network. We then apply the embedding operators.

\item The link prediction heuristic scores JC, CN, AA and PA are applied individually on each layer of the multiplex networks. We then average the scores, as JC-av, CN-av, AA-av, and PA-av.
\end{itemize}

From the outputs of the embedding operators and heuristic scores, we feed and train a binary classifier and then test it on the 30\% of test edges that have been removed initially. The binary classifier is a logistic regressor.

The evaluation metrics for link prediction is ROC-AUC as it is commonly used for embedding evaluation on link prediction and to validate network embedding \cite{zhang2018scalable, grover2016node2vec}. The ROC-AUC is computed as the area under the ROC curve, which plots the true positive rate (TPR) against the false positive rate (FPR) at various threshold settings. An AUC value of $1$ represent a model that classifies perfectly the samples. 

\subsubsection*{Network reconstruction}
\label{sec_network_rec}

Network reconstruction is another approach to evaluate network embedding methods \cite{wang2016structural, bagavathi2018multi, goyal2018graph}. In this case, the goal is to quantify the amount of topological information captured by the embedding methods. This is equivalent to predict if we can go back from the embedding to the original adjacency matrix of each layer of the multiplex network.

\begin{figure}[t]
\begin{center}
\includegraphics[width=1\linewidth]{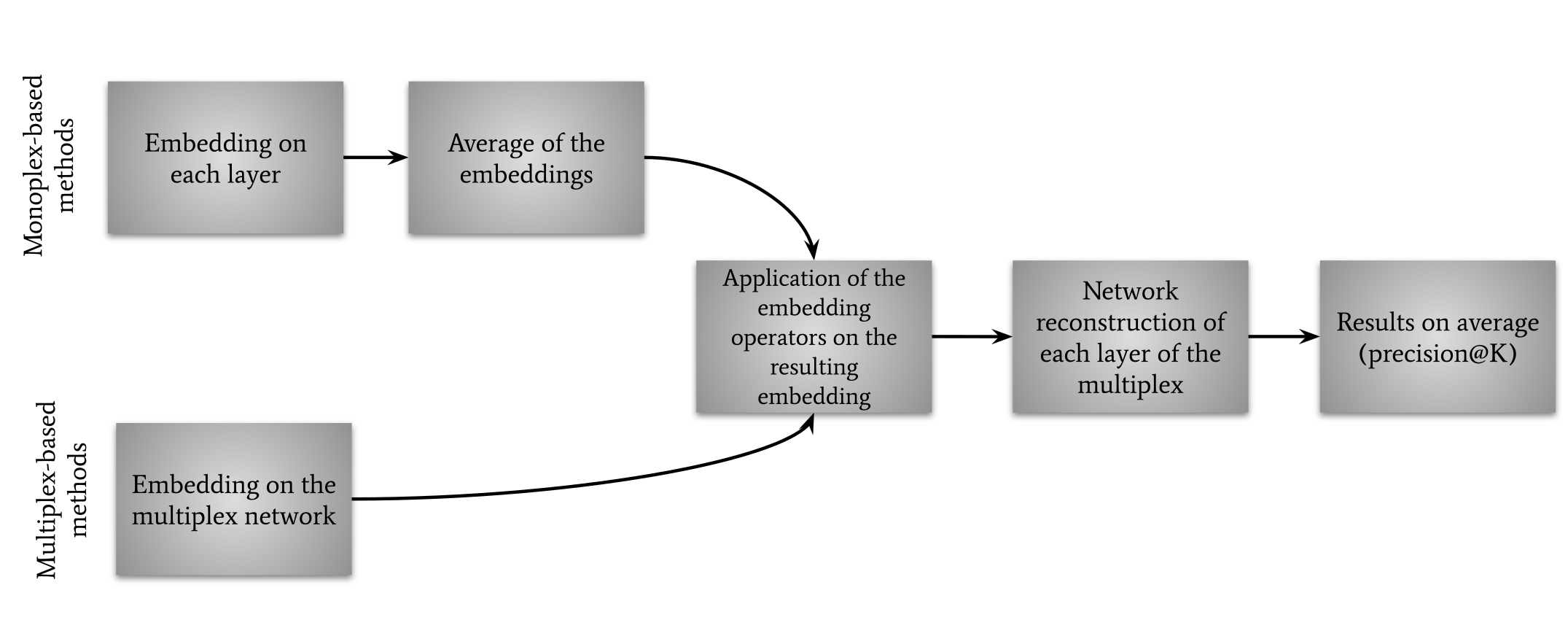} 
\end{center}
\caption{General approach for network reconstruction on multiplex networks: (top) for monoplex-based methods, embed each layer with the given method, then average it~; (bottom) for multiplex-based methods, apply the specific embedding method to the network. Embedding operators are then applied to monoplex- and multiplex-based method embeddings. The three types of methods are finally evalutated for network embedding using a binary classifier and a precision@K score is computed.}
\label{net_rec}
\end{figure}

Theoretically, to reconstruct the networks, one would need to apply link prediction to every possible edge in the graphs. This is however in practice not scalable to large graphs. Indeed, it would correspond to $n(n-1)/2$ potential edges to classify (for undirected networks of $n$ nodes without self-loops). In addition, the networks in our study are sparse, with much more false (absent) than true (present) edges, leading to large class imbalance. In this context, ROC-AUC can be misleading, as large changes in the ROC Curve or ROC-AUC score can be caused by a small number of correct or incorrect predictions \cite{fernandez2018learning}. In order to account for class imbalance, we used the precision@$K$ \cite{wang2016structural}. This evaluation metric is based on the sorting in descending order of all predictions and consider the first $K$ best predictions  to evaluate how many true edges (the minority class) are predicted correctly by the binary classifier. 
From the outputs of the embedding operators, we perform network reconstruction by training a binary classifier on a subset of the original networks (Figure \ref{net_rec}). We choose a subset of 95\% of the edge pairs from the original adjacency matrix of each layer for the smaller multiplex networks (CKM, LAZEGA and C.ELE) to construct the training graph. As the class imbalance increases with the number of nodes and sparsity of the networks, we choose smaller subsets for the largest networks, respectively 5\% of edges for the ARXIV network and 2.5\% for the other networks, as in previous publications \cite{goyal2018graph, wang2016structural}. For each layer, $K$ is defined as the maximum of true edges in this subset of edge pairs. We use a Random Forest algorithm as a binary classifier for network reconstruction, as it is known to be less sensitive to class imbalance \cite{more2017review}. In network reconstruction, the results correspond to the training phase of the classifier, there is no test phase.

\subsection{Evaluation of multiplex-heterogeneous network embedding}

\subsubsection{Multiplex-heterogeneous network datasets}
\label{ssec:ref_MH}

\begin{itemize}
\itemsep0em 

\item \textbf{Gene-disease multiplex-heterogeneous network}: We use the two multiplex networks presented in the previous sections: the disease (DIS) and molecular multiplex networks (MOL) (Table \ref{sample-table}).
In addition, we extracted the curated gene-disease bipartite network from the DisGeNET database \cite{pinero2020disgenet} in order to connect the two multiplex networks. This bipartite interaction network contains $75445$ interactions between $5188$ diseases and $9179$ genes. We obtain a multiplex-heterogeneous network, as represented in Figure \ref{net_All}C.

\item \textbf{Drug-target multiplex-heterogeneous network}: We use the molecular multiplex network (MOL) from the previous multiplex-heterogeneous network. We constructed the following 3-layers drug multiplex network: (i) the first layer ($2795$ edges, $877$ nodes) has been extracted from Bionetdata (\url{https://rdrr.io/cran/bionetdata/man/DD.chem.data.html}) and the edges correspond to Tanimoto chemical similarities between drugs if superior to 0.6, (ii) the second layer ($678$ edges, $362$ nodes) comes from \cite{cheng2019network} and the edges are based on drug combinations as reported in clinical data, (iii) the third layer ($13397$ edges, $658$ nodes) is the adverse drug–drug interactions network available in \cite{cheng2019network}. The drug-target bipartite network has been extracted from the same publication \cite{cheng2019network}, and contains $15030$ bipartite interactions between $4412$ drugs and $2255$ protein targets.


\end{itemize}

\subsubsection{Evaluation task}

We validate the multiplex-heterogeneous network embedding using link prediction. We remove randomly 30\% of the edges but only from the bipartite interactions to obtain a training graph. We then train a Random Forest on the training graph, and test on the 30\% removed edges. Based on the multiplex-heterogeneous networks described previously, the idea behind this evaluation is to test if we can predict gene-disease and drug-gene links. Comparisons with other approaches are not possible as, to our knowledge, no existing multiplex-heterogeneous network embedding method are currently available in the literature.

\subsection{Case study: discovery of new gene-disease associations}

\subsubsection*{Link prediction}

Our aim for this case-study is to predict new gene-disease links. We thereby applied MultiVERSE on the full gene-disease multiplex-heterogeneous network without removing any edges, and trained a binary classifier (Random Forest) using edges from the bipartite interactions. Then, we test all possible gene-disease edges that are not in the original bipartite interactions and involve Progeria and Xeroderma pigmentosum VII disease nodes. Finally, we select the top 5 new gene-disease associations for each disease.

\subsubsection*{Clustering}
\label{ssec:clustering_MH}

We also applied MultiVERSE to the gene-disease multiplex-heterogeneous gene-disease network, followed by spherical K-means \cite{buchta2012spherical} to cluster the vector representations of nodes. Spherical K-means clustering is well-adapted to high-dimensional clustering \cite{zhong2005efficient}. We define the number of clusters for spherical K-means to $500$, in order to obtain cluster sizes that can be analysed from a biological point of view.

\section{Results}
 \label{sec:results}
 
\subsection{Evaluation results for multiplex network embeddings}
\label{ssec:results_M}

\subsubsection{Link prediction}

We evaluate the performance of the different methods (link prediction heuristics and network embedding) on the task of link prediction applied to the set of multiplex networks. First, we can observe that the heuristics are not efficient for link prediction, with ROC-AUC only slightly better than random classification (Table \ref{linkpred_table}). 

The methods based on embedding always perform better than the heuristic baselines. In addition, the ROC-AUC is in most of the cases higher when the models take into account the multiplex network structure rather than the monoplex-average, as observed in \cite{zhang2018scalable}. For instance, using the Hadamard operator, the ROC-AUC average over all the networks of the three monoplex-average approaches (node2vec-av, deepwalk-av, LINE-av) is $0.8025$, whereas the average of the three multiplex-based approaches (Ohmnet, MNE, Multi-node2vec) is $0.8381$. The ROC-AUC score average of MultiVERSE in this context is $0.9011$.
Nevertheless, node2vec-av and deepwalk-av perform very well and even outperform multiplex-based approaches on various scenarios, for instance link prediction on the C.ELE and ARXIV networks.

MultiVERSE combined with the Hadamard operator outperforms the other methods for all the tested networks but CKM. In addition, MultiVERSE is the best approach when combined with three out of five operators (Hadamard, Average, Cosine). These results suggest that RWR-M is able to better capture the topological features of the networks under study.



\begin{table*}[t!]
\centering
\resizebox{\textwidth}{!}{%
\begin{tabular}{l l l l l l l l l}
\thead{ Operators} & \thead{Method} & \thead{CKM} & \thead{LAZEGA} & \thead{C.ELE} & \thead{ARXIV} &  \thead{DIS} & \thead{HOMO} & \thead{MOL} \\ \hlineB{2.5}
\multirow{4}{*}{Link prediction heuristics} 
 & CN-av & 0.4944 & 0.6122 & 0.5548 & 0.5089 & 0.5097 & 0.5113 & 0.5408 \\ 
 & AA-av & 0.4972 & 0.6105 & 0.549 & 0.5081 & \underline{0.5428} & 0.5112 & 0.5404 \\ 
 & JC-av & 0.4911 & 0.523 & 0.5424 & 0.5113 & 0.5425 & 0.5113 & \underline{0.5433} \\ 
 & PA-av & \underline{0.5474} & \underline{0.6794} & \underline{0.5634} & \underline{0.5139} & 0.496 & \underline{0.5185} & 0.5278 \\ \hline
 
\multirow{6}{*}{Hadamard} 
& node2vec-av & 0.7908 & 0.6372 & 0.8552 & 0.9775 & 0.9093 & 0.8638 & 0.8753 \\ 
 & deepwalk-av & 0.7467 & 0.6301 & 0.8574 & 0.9776 & 0.9107 & 0.8638 & 0.8763 \\ 
 & LINE-av & 0.5073 & 0.4986 & 0.5447 & 0.8525 & 0.9013 & 0.8852 & 0.8918 \\ 
 \cmidrule(lr){2-9}
 & Ohmnet & 0.7465 & 0.7981 & 0.833 & 0.9605 & 0.9333 & 0.9055 & 0.8613 \\ 
 & MNE & 0.5756 & 0.6356 & 0.794 & 0.9439 & 0.9099 & 0.8313 & 0.8736 \\ 
 & Multi-node2vec & 0.8182 & 0.7884 & 0.8375 & 0.9581 & 0.8528 & 0.8592 & 0.8835 \\ 
 \cmidrule(lr){2-9}
 & MultiVERSE & \underline{0.8177} & \underline{\textbf{0.8269}} & \underline{\textbf{0.8866}} & \underline{\textbf{0.9937}} & \underline{\textbf{0.9401}} & \underline{\textbf{0.917}} & \underline{\textbf{0.9259}} \\ \hline
 
\multirow{6}{*}{Weighted-L1}
& node2vec-av & 0.7532 & 0.737 & \underline{0.8673} & 0.9738 & 0.885 & 0.6984 & 0.7976 \\ 
 & deepwalk-av & 0.7226 & 0.7094 & 0.8635 & \underline{0.9751} & \underline{0.8888} & 0.7142 & 0.8089 \\ 
 & LINE-av & 0.6091 & 0.5776 & 0.6192 & 0.7539 & 0.8586 & 0.7439 & 0.7792 \\ 
 \cmidrule(lr){2-9}
 & Ohmnet & 0.7421 & 0.7849 & 0.8128 & 0.8488 & 0.8503 & 0.7007 & 0.6983 \\ 
 & MNE & 0.6289 & 0.6523 & 0.8019 & 0.7805 & 0.8313 & 0.7619 & 0.8182 \\ 
 & Multi-node2vec & \underline{0.8611} & \underline{0.8089} & 0.8261 & 0.9659 & 0.8628 & \underline{0.8472} & \underline{0.8997} \\
 \cmidrule(lr){2-9} 
 & MultiVERSE & 0.7043 & 0.7789 & 0.7516 & 0.8647 & 0.7754 & 0.683 & 0.7273 \\ \hline

\multirow{6}{*}{Weighted-L2} 
 & node2vec-av & 0.7556 & 0.6851 & \underline{0.8691} & 0.9743 & 0.8867 & 0.7048 & 0.8028 \\ 
 & deepwalk-av & 0.7221 & 0.6904 & 0.864 & \underline{0.9771} & \underline{0.8891} & 0.7145 & 0.813 \\ 
 & LINE-av & 0.5851 & 0.5756 & 0.6275 & 0.7609 & 0.8621 & 0.7429 & 0.7835 \\ 
 \cmidrule(lr){2-9}
 & Ohmnet & 0.7505 & 0.7788 & 0.8166 & 0.8439 & 0.8599 & 0.7041 & 0.6992 \\ 
 & MNE & 0.601 & 0.5397 & 0.7999 & 0.7815 & 0.8333 & 0.7483 & 0.8122 \\ 
 & Multi-node2vec & \underline{\textbf{0.8637}} & \underline{0.8091} & 0.8282 & 0.968 & 0.8675 & \underline{0.8525} & \underline{0.9004} \\
 \cmidrule(lr){2-9} 
 & MultiVERSE & 0.7125 & 0.7801 & 0.7441 & 0.8661 & 0.7808 & 0.6918 & 0.7475 \\ \hline
 
\multirow{6}{*}{Average} 
 & node2vec-av & 0.59 & 0.6596 & 0.6842 & 0.6615 & 0.8256 & 0.8308 & 0.777 \\ 
 & deepwalk-av & 0.5954 & 0.657 & 0.6784 & 0.6582 & 0.8267 & 0.8307 & 0.7737 \\ 
 & LINE-av & 0.5465 & 0.6581 & 0.6699 & 0.6465 & 0.8477 & 0.8653 & \underline{0.8276} \\ 
 \cmidrule(lr){2-9}
 & Ohmnet & 0.5764 & 0.656 & 0.7334 & \underline{0.6772} & 0.8533 & 0.8825 & 0.7962 \\ 
 & MNE & 0.5882 & 0.6615 & 0.7028 & 0.6723 & 0.8242 & 0.8024 & 0.783 \\ 
 & Multi-node2vec & 0.5571 & 0.6584 & 0.7365 & 0.6657 & 0.8222 & 0.8216 & 0.7589 \\ 
 \cmidrule(lr){2-9}
 & MultiVERSE & \underline{0.5963} & \underline{0.6728} & \underline{0.7438} & 0.6752 & \underline{0.8586} & 0.8643 & 0.812 \\ \hline
 
\multirow{6}{*}{Cosine} 
 & node2vec-av & 0.7805 & 0.7335 & 0.8515 & 0.9711 & 0.8643 & 0.7368 & 0.8105 \\ 
 & deepwalk-av & 0.7465 & 0.7066 & 0.8416 & 0.9724 & 0.8667 & 0.7512 & 0.8079 \\ 
 & LINE-av & 0.545 & 0.5126 & 0.5477 & 0.8198 & 0.7409 & 0.6745 & 0.816 \\ 
  \cmidrule(lr){2-9}
 & Ohmnet & 0.7898 & 0.7352 & 0.8094 & 0.9642 & 0.859 & 0.7829 & 0.7909 \\ 
 & MNE & 0.6203 & 0.6506 & 0.7877 & 0.8951 & 0.8347 & 0.6474 & 0.8102 \\ 
 & Multi-node2vec & \underline{0.8532} & 0.7931 & 0.7815 & 0.9435 & 0.7151 & 0.8477 & 0.8884 \\ 
  \cmidrule(lr){2-9}
 & MultiVERSE & 0.8148 & \underline{0.8171} & \underline{0.8719} & \underline{0.9909} & \underline{0.8775} & \underline{0.8776} & \underline{0.9103} \\
\end{tabular}
}
\caption{ROC-AUC scores for link prediction on the 7 reference multiplex networks, for link prediction heuristics (CN-av, AA-av, JC-av, PA-av) and network embedding methods combined with different operators (Hadamard, Weighted-L1, Weighted-L2, Average, and Cosine). For each multiplex network, the best score is in bold; for each operator, the best scores are underlined. Overall, the MultiVERSE algorithm combined with the Hadamard shows the best scores.}
\label{linkpred_table}
\end{table*}

\subsubsection{Network reconstruction}

We next evaluate the performances of the different embedding methods on the task of network reconstruction applied to multiplex networks. As described in section \ref{sec_network_rec}, we now rely on the evaluation metric, precision@$K$. The experimental results are shown in Table \ref{netrec_table}. 

On one hand, for the small networks (i.e., CKM, LAZEGA and C.ELE), the best precision is achieved with LINE-av in combination with any of the operators but Cosine. In particular, LINE-av obtains a perfect score for the CKM network using the Weighted-L2 or Hadamard operators. MNE is in second position with more than 99\% of precision using the Weighted-L1 or Weighted-L2 operators. LINE-av also presents good performances for the C.ELE network with a precision of $93.67\%$ using the Weighted-L2 operator, almost $20\%$ higher than the second best method on this network (Multi-node2vec with a score of $0.7568$ using the  Weighted-L2 operator).

\begin{table*}[t!]
\centering
\resizebox{\textwidth}{!}{%
\begin{tabular}{l l l l l l l l l}
\thead{Operators} & \thead{Method} & \thead{CKM\\(95\%)} & \thead{LAZEGA\\(95\%)} & \thead{C.ELE\\(95\%)} & \thead{ARXIV\\(5\%)} &  \thead{DIS\\(2,5\%)} & \thead{HOMO\\(2,5\%)} & \thead{MOL\\(2.5\%)} \\ \hlineB{2.5}

\multirow{6}{*}{Hadamard} 
 & node2vec-av & 0.6764 & 0.9174 & 0.4526 & 0.8207 & 0.5578 & 0.7599 & 0.2989 \\ 
 & deepwalk-av & 0.6564 & 0.9351 & 0.4416 & 0.7886 & 0.5486 & 0.7636 & 0.3164 \\ 
 & LINE-av & \underline{\textbf{1.0}} & \underline{0.9924} & \underline{0.8924} & 0.8204 & 0.4955 & 0.5191 & 0.4006 \\ 
     \cmidrule(lr){2-9} 
 & Ohmnet & 0.7842 & 0.8334 & 0.5329 & 0.9156 & 0.4811 & 0.6979 & 0.2591 \\ 
 & MNE & 0.9505 & 0.9094 & 0.2728 & 0.7891 & 0.4218 & 0.3641 & 0.1316 \\ 
 & Multi-node2vec & 0.8352 & 0.8811 & 0.6875 & 0.8605 & 0.6063 & 0.7584 & 0.3123 \\ 
     \cmidrule(lr){2-9} 
 & MultiVERSE & 0.9687 & 0.9695 & 0.7436 & \underline{0.9015} & \underline{0.6734} & \underline{\textbf{0.8729}} & \underline{0.3674} \\ \hline
 
\multirow{6}{*}{Weighted-L1}
& node2vec-av & 0.5923 & 0.9494 & 0.5129 & 0.6922 & 0.5859 & 0.8123 & 0.3194 \\ 
 & deepwalk-av & 0.5791 & 0.9784 & 0.4896 & 0.6878 & 0.5921 & 0.7984 & 0.3206 \\ 
 & LINE-av & \underline{0.9985} & \underline{0.9953} & \underline{0.9229} & 0.7837 & 0.4921 & 0.6839 & 0.3586 \\ 
     \cmidrule(lr){2-9} 
 & Ohmnet & 0.7355 & 0.8581 & 0.5785 & 0.8771 & 0.6025 & \underline{0.8019} & \underline{0.3769} \\ 
 & MNE & 0.9926 & 0.975 & 0.4722 & 0.8593 & 0.4377 & 0.5241 & 0.1861 \\ 
 & Multi-node2vec & 0.8636 & 0.9235 & 0.7379 & 0.7684 & 0.6356 & 0.7649 & 0.2671 \\ 
     \cmidrule(lr){2-9} 
 & MultiVERSE & 0.8545 & 0.9638 & 0.7444 & \underline{0.8705} & \underline{0.6678} & 0.7913 & 0.3559 \\ \hline
 
\multirow{6}{*}{Weighted-L2} 
& node2vec-av & 0.5886 & 0.9436 & 0.5097 & 0.6983 & 0.5953 & \underline{0.8193} & 0.352 \\ 
 & deepwalk-av & 0.5829 & 0.9672 & 0.5146 & 0.6877 & 0.5857 & 0.805 & 0.3233 \\ 
 & LINE-av & \underline{\textbf{1.0}} & \underline{\textbf{0.9962}} & \underline{\textbf{0.9367}} & 0.7749 & 0.4945 & 0.6697 & 0.392 \\ 
      \cmidrule(lr){2-9} 
 & Ohmnet & 0.7418 & 0.8687 & 0.5724 & 0.8694 & 0.6209 & 0.8143 & \underline{0.3701} \\ 
 & MNE & 0.9926 & 0.9764 & 0.4646 & \underline{0.8818} & 0.4351 & 0.5529 & 0.176 \\ 
 & Multi-node2vec & 0.8644 & 0.93 & 0.7568 & 0.7548 & 0.6361 & 0.7896 & 0.2922 \\ 
      \cmidrule(lr){2-9} 
 & MultiVERSE & 0.8653 & 0.969 & 0.754 & 0.8776 & \underline{\textbf{0.6784}} & 0.7876 & \underline{0.3701} \\ \hline

\multirow{6}{*}{Average} 
& node2vec-av & 0.8408 & 0.917 & 0.4817 & 0.889 & 0.5587 & 0.6809 & 0.2686 \\ 
 & deepwalk-av & 0.8331 & 0.9379 & 0.501 & 0.8853 & 0.5318 & 0.6714 & 0.2795 \\ 
 & LINE-av & \underline{0.9855} & \underline{0.9382} & \underline{0.7103} & 0.8725 & 0.5093 & 0.5677 & 0.3244 \\
      \cmidrule(lr){2-9} 
 & Ohmnet & 0.9412 & 0.8287 & 0.5825 & 0.906 & 0.4989 & 0.6551 & 0.2887 \\ 
 & MNE & 0.9179 & 0.9151 & 0.2966 & 0.7146 & 0.4175 & 0.352 & 0.1444 \\ 
 & Multi-node2vec & 0.9767 & 0.8937 & 0.6726 & 0.9498 & 0.6243 & 0.6216 & 0.2901 \\ 
      \cmidrule(lr){2-9} 
 & MultiVERSE & 0.978 & 0.9059 & 0.5326 & \underline{\textbf{0.9758}} & \underline{0.6316} & \underline{0.7204} & \underline{\textbf{0.4143}} \\ \hline

\multirow{6}{*}{Cosine} 
& node2vec-av & 0.5103 & 0.4936 & 0.18 & 0.2537 & 0.1825 & 0.116 & 0.0441 \\ 
 & deepwalk-av & 0.4807 & 0.4776 & 0.1741 & \underline{0.2835} & 0.1854 & 0.1036 & 0.0462 \\ 
 & LINE-av & 0.3291 & 0.4974 & \underline{0.1867} & 0.2638 & \underline{0.2384} & 0.1476 & 0.0454 \\ 
       \cmidrule(lr){2-9} 
 & Ohmnet & 0.5696 & 0.509 & 0.1718 & 0.2655 & 0.1984 & 0.1311 & 0.044 \\ 
 & MNE & 0.3169 & 0.4536 & 0.1768 & 0.2445 & 0.1957 & \underline{0.1667} & 0.044 \\ 
 & Multi-node2vec & 0.5127 & \underline{0.52} & 0.186 & 0.273 & 0.195 & 0.1032 & 0.0461 \\
       \cmidrule(lr){2-9} 
 & MultiVERSE & \underline{0.6395} & 0.5026 & 0.1818 & 0.254 & 0.1983 & 0.1522 & \underline{0.0474} \\
\end{tabular}
}
\caption{precision@$K$ scores for network reconstruction on the 7 reference multiplex networks, for the network embedding methods combined with different embeddings operators (Hadamard, Weighted-L1, Weighted-L2, Average, and Cosine). For each multiplex network, the best score is in bold; for each operator, the best score is underlined. The percentage of edges used for the reconstruction  is indicated in parenthesis under the name of the network. In the case of large networks  (DIS, ARXIV, HOMO and MOL) MultiVERSE achieves the best performance in combination with different operators.}
\label{netrec_table}

\end{table*}

On the other hand, we can group together the results obtained for the large networks (DIS, ARXIV, HOMO and MOL). In this case, MultiVERSE achieves the best performance in combination with different operators. Large networks are sparse, leading to high class imbalance (\ref{sec_network_rec}). Still, MultiVERSE achieves a good score for the HOMO and DIS networks, with precision@$K$ of $0.8729$ and $0.6784$, respectively. The precision obtained on the molecular network (MOL) is the lowest, with a precision@$K$ of $0.4143$. The complexity of the task is possibly higher as the number of nodes and class imbalance increase.

Overall, the lowest scores are obtained by MNE and, in general, the Cosine operator performs poorly for all methods. The network reconstruction process is a complex task, and the performance depends on the size and density of the different layers composing the multiplex network. Nevertheless, MultiVERSE obtains good results for most of the networks without any processing of the imbalanced data.

\subsection{Evaluation results for multiplex-heterogeneous network embedding}
\label{ssec:results_MH}


The task of link prediction on multiplex-heterogeneous networks is applied to MultiVERSE only, as to our knowledge no other methods exist for the embedding of multiple nodes from multiplex-heterogeneous networks. MultiVERSE has a score of ROC-AUC superior to 0.9 with the Hadamard and Average operators (Table \ref{results_MH}), meaning that the method can predict with high precision the gene-disease and drug-disease links from the corresponding multiplex-heterogeneous networks.

\begin{table}[t]
\begin{center}


\begin{tabular}{c c c}
Operators & Gene-Disease Bipartite & Drug-target Bipartite \\ \hlineB{2.5}
Hadamard &  \textbf{0.95} & \textbf{0.9701} \\ 
Weighted-L1 &  0.7962 & 0.8057 \\ 
Weighted-L2 & 0.7951 & 0.8055 \\ 
Average & \textbf{0.9603} & \textbf{0.9703}\\ 
Cosine & 0.7765 & 0.8338 \\ 
\end{tabular}
\end{center}
\caption{ROC-AUC scores for link prediction using MultiVERSE on 2 multiplex-heterogeneous reference networks. Link predictions are computed for the bipartite interactions of the multiplex-heterogeneous networks. The scores higher than 0.9 are highlighted in bold.}
\label{results_MH}
\end{table}

\subsection{Case study results: discovery of new gene-disease associations}

\subsubsection{Discovery of new gene-disease associations with link prediction}

The results of the evaluations on multiplex-heterogeneous network link prediction show that MultiVERSE combined with the Hadamard and Average operators reach ROC-AUC scores superior to 0.9 (Table \ref{results_MH}). We here investigate in detail the top 5 new gene-disease associations predicted by MultiVERSE combined with these operators for Hutchinson-Gilford Progeria Syndrome (HGPS) and Xeroderma  pigmentosum VII (Table \ref{results_linkpred_application}) .


\begin{table}[ht]
\centering
\begin{tabular}{cc|cc}
\multicolumn{2}{c|}{\textbf{HGPS}}  & \multicolumn{2}{c}{\textbf{Xeroderma p. VII}} \\
\thead{Average}  & \thead{Hadamard}  & \thead{Average}            & \thead{Hadamard}          \\ \hlineB{2.5}
\textit{NOS2}             & \textit{POT1}            & \textit{TNF}          & \textit{TNF}                      \\
\textit{IL6}              & \textit{TERF1}             & \textit{SOD2}      & \textit{VCAM1}                     \\
\textit{TNF}              & \textit{EEF1A1}           & \textit{IL6}        & \textit{NUP62}                     \\
\textit{SOD1}             & \textit{TERF2}          & \textit{TP53}         & \textit{ERCC2}                     \\
\textit{SOD2}             & \textit{TERF2IP}          & \textit{FN1}        & \textit{MCC}                    
\end{tabular}
\caption{Top 5 predictions of new gene-disease associations for HGPS and Xeroderma pigmentosum VII by MultiVERSE combined with Average and Hadamard operators.}
\label{results_linkpred_application}
\end{table}

\subsubsection*{Hutchinson-Gilford Progeria Syndrome}

Hutchinson-Gilford Progeria Syndrome (HGPS) is a rare premature aging genetic disease characterized by postnatal growth retardation, midface hypoplasia, micrognathia, premature atherosclerosis, coronary artery disease, lipodystrophy, alopecia and generalized osteodysplasia \cite{de2003lamin}. HGPS is caused by mutations in the \textit{LMNA} genes that cause the production of a toxic form of the Lamin A protein called Progerin.

MultiVERSE top predictions reveal interesting candidate genes (Table \ref{results_linkpred_application}). In particular, \textit{NOS2} encodes a nitric oxide synthase expressed in liver. It has been associated with longevity \cite{montesanto2013common}.
\textit{TNF} is a member of the tumor necrosis factor superfamily, and produces a multifunctional proinflammatory cytokine. \textit{TNF} is also known to be involved in aging \cite{davizon2019tnf} and has been previously linked to Progeria \cite{osorio2012nuclear}.
\textit{TERF1} and \textit{TERF2} both encode telomere-binding proteins and \textit{TERF2IP} encodes a protein that is part of a complex involved in telomere length regulation. HGPS patients show increased activation of DNA damage signalling at telomeres associated to reduced telomere length \cite{decker2009telomere}. In addition, it has been reported DNA damage accumulation and \textit{TRF2} degradation in atypical Werner syndrome (adult Progeria) fibroblasts with \textit{LMNA} mutations \cite{saha2013dna}. 
\textit{POT1} also produces a telomeric protein that has been linked to the 
Werner syndrome \cite{sowd2008mechanism}, the maintenance of haematopoeitic stem cell activity during aging \cite{hosokawa2017telomere} and cellular senescence \cite{li2019seryl}.
\textit{IL6} encodes a cytokine involved in inflammation, which have also been linked to aging \cite{maggio2006interleukin}. 
Finally,\textit{SOD1} and \textit{SOD2} are members of the superoxide dismutase multigene family that destroy free superoxide radicals. They both  have been associated to aging and cellular senescence \cite{zhang2017new, velarde2012mitochondrial}.


\subsubsection*{Xeroderma pigmentosum VII}

Xeroderma Pigmentosum (XP) is characterized by extreme sensitivity to sunlight, resulting in sunburns, pigment changes in the skin and a highly elevated incidence of melanoma. It is a genetically heterogeneous autosomal recessive disorder. Several XP types exist, and the MeSH term Xeroderma pigmentosum VII corresponds to the group G, caused by mutations in the \textit{ERCC5} gene, and with symptoms that overlap Cockayne syndrome  \cite{kraemer1987xeroderma, vermeulen1993xeroderma}.
 
MultiVERSE identified various candidates for this disease (see Table \ref{results_linkpred_application}), we detail here the most interesting ones.
\textit{TNF} has been related to XP \cite{capulas2000ultraviolet} and skin tumour development \cite{arnott2004expression}.
\textit{IL6} is involved in melanoma, one of the major phenotypes of XP \cite{lu1992interleukin}.
\textit{SOD2}, also predicted as candidate for HGPS, have been recently associated to melanoma \cite{yuan2020braf}.
\textit{TP53} is a tumor suppressor implicated in many cancers, in particular melanomas \cite{giglia2003tp53}. It has also been associated to XP \cite{sarasin2019familial}.
\textit{VCAM1} encodes the Vascular Cell Adhesion Molecule, associated to melanoma \cite{klemke2007high}, and
\textit{NUP62} encodes the Nuclear pore glycoprotein p62, and involved in the regulation of squamous cell carcinoma proliferation \cite{hazawa2018rock}.
\textit{ERCC2} produces the  \textit{XPD} protein, mutat in XP group D \cite{taylor1997xeroderma}.



\subsubsection{Discovery of new gene-disease associations with clustering}

Another illustration of the advantages of multiplex-heterogeneous network embedding is clustering. We identify clusters with K-means (see section \ref{ssec:clustering_MH}), and focus more particularly on the clusters containing HGPS and Xeroderma pigmentosum VII disease nodes. 
Clustering is particularly interesting as it can be applied directly on the embeddings without supervised training. In addition, it has been shown that clustering  from embeddings outperforms the other methods for the detection of biological communities \cite{nelson2019embed}.

\subsubsection*{Cluster containing the HGPS disease node}

\begin{figure}[t]
\begin{center}
\includegraphics[width=1\hsize]{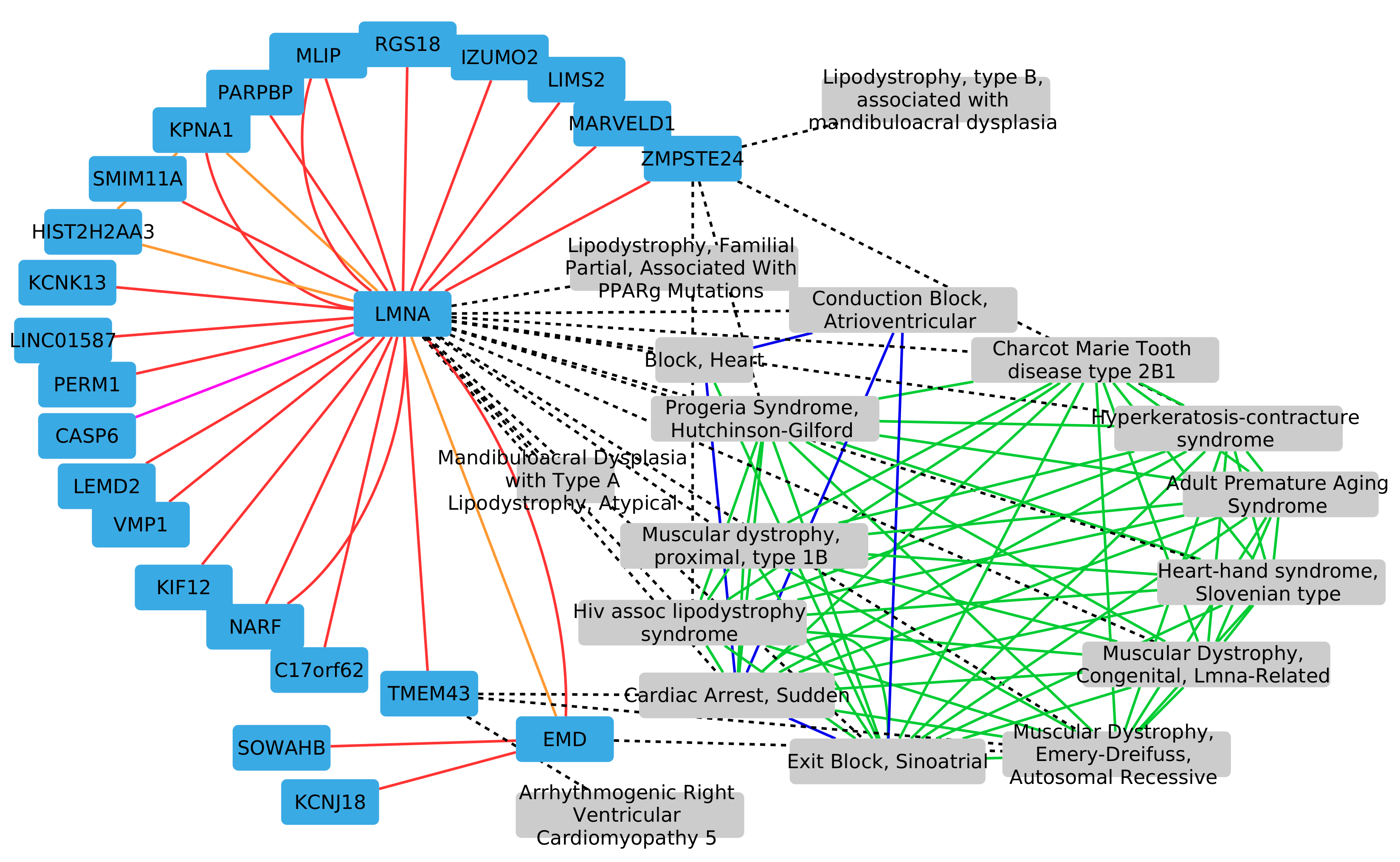} 
\end{center}
\caption{Cluster containing the HGPS disease node. Disease-Disease edges from the disease multiplex network are represented in green (shared symptoms) and blue (CTD projection).  Gene-Gene edges from the molecular multiplex network are represented in pink (Reactome pathways), red (protein-protein interactions) and orange (molecular complexes). Gene-Disease bipartite interactions are represented with black dashed lines. }
\label{LMNA_cluster}
\end{figure}

The cluster containing the HGPS disease node (see Figure \ref{LMNA_cluster}) contains the \textit{LMNA} node. \textit{LMNA} mutations have been observed in many diseases that also belong to the identified cluster, including the Heart-hand syndrome (Slovenian type)\cite{renou2008heart}, lipodystrophy  associated with mandibuloacral dysplasia \cite{agarwal2008severe}, the Charcot-Marie-Tooth disease, type 2B1 \cite{sinha2014progeria}, \textit{LMNA}-related muscular distrophy, and different cardiac diseases caused by \textit{LMNA} mutations \cite{van2007high}. 

We also analysed the cluster's genes annotations with g:Profiler (\cite{raudvere2019g}, default parameters). 
We found significant enrichments in several annotations related to cell nuclear organization. One of the most significant enrichments is nuclear envelope, involving the following genes: \textit{EMD}, \textit{LEMD2}, \textit{LMNA}, \textit{KPNA1}, \textit{MLIP}, \textit{TMEM43}, and \textit{ZMPSTE24}. HGPS is a disorder of the nuclear envelope \cite{worman2010diseases}.


\subsubsection*{Cluster containing the Xeroderma pigmentosum VII disease node}

\begin{figure}[t]
\begin{center}
\includegraphics[width=0.95\hsize]{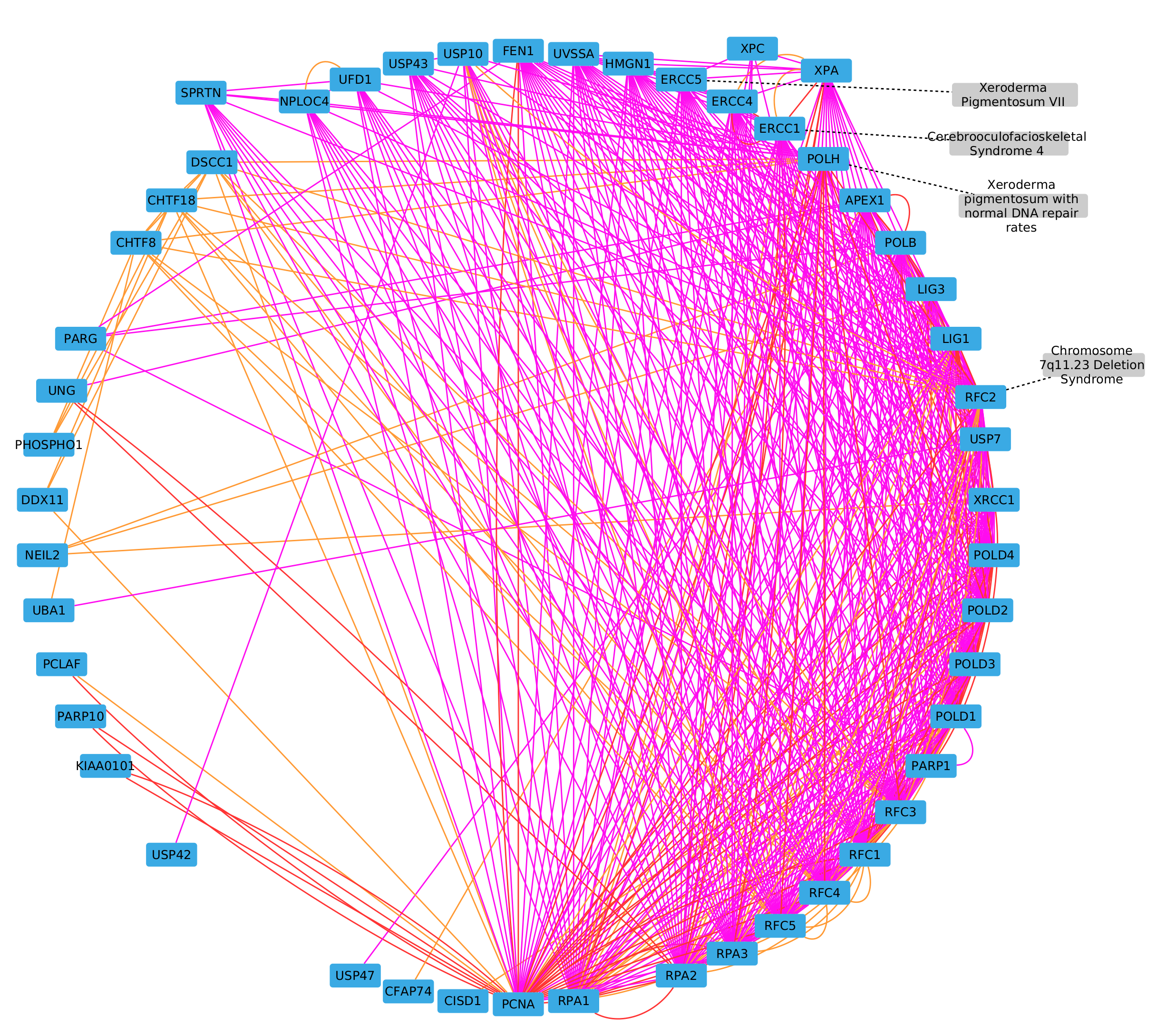} 
\end{center}
\caption{Cluster containing the Xeroderma pigmentosum VII disease node.  Gene-Gene edges from the molecular multiplex network are represented in pink (Reactome pathways), red (protein-protein interactions) and orange (molecular complexes). Gene-Disease bipartite interactions are represented with black dashed lines. }
\label{Xeroderma_cluster}
\end{figure}

We also analysed the cluster containing the Xeroderma pigmentosum VII disease node (Figure \ref{Xeroderma_cluster}). The cluster contains different diseases, including  Xeroderma pigmentosum with normal DNA repair rates, and Cerebro-oculo-facio-skeletal syndrome 4, which is also a nuclear-excision repair disorder \cite{graham2001cerebro}. 


Several genes known for their implication in XP are present in the cluster, such as \textit{ERCC1}, \textit{ERCC4}, \textit{ERCC5}, \textit{XPA} and \textit{XPC} \cite{sugasawa2008xeroderma}. Using the complete list of genes in the cluster as an input for g:Profiler (\cite{raudvere2019g}, default parameters), we identified several significantly enriched annotations. Among them, we can cite nucleotide-excision DNA repair, defective DNA repair after ultraviolet radiation damage or response to  ultraviolet radiation. XP patients show important impairments in these biological processes \cite{kraemer1987xeroderma}. 
\textit{SPRTN} is another gene of interest. It encodes a metalloprotease that repairs DNA-protein crosslinks. \textit{SPRTN} does not share interactions with genes known to be mutated in XP, but has been shown to be involved in UV sensibilization and cancer \cite{hiom2014sprtn}.


\section{Discussion and conclusion}

We present in this study MultiVERSE, a new approach for multiplex and multiplex-heterogeneous network embedding. MultiVERSE is fully parallelized and scalable, even if the current implementation requires the generation of dense matrices, which can raise memory issues when dealing with very large networks.

For multiplex network embedding, we compared MultiVERSE with state-of-the-art methods using link prediction and network reconstruction. We show that MultiVERSE outperforms various methods specifically developed for multiplex network embedding. As RWR-M applies a random walk in pseudo-infinite time, it might allow MultiVERSE to effectively capture node properties and a better representation of the topological structure of the multiplex network. 

A natural extension of this work would be to consider multiplex networks composed of both directed and undirected layers. In a biological context, this would allow considering metabolic and signalling pathways networks into a multiplex structure without losing the information about the information flow. In addition, for the optimization phase, we set a neighborhood parameter $N_{max}$ that depends on the size of the network. A potential improvement could be to develop an adaptive version of the parameter $N_{max}$ that would depend on node topological properties.

For multiplex-heterogeneous network embedding, MultiVERSE allows the embedding of different types of nodes. We demonstrate its efficiency for link prediction and illustrates its usefulness for the study of gene-disease associations. We here limited the multiplex-heterogeneous network to two multiplex and one bipartite network. Another natural extension of our work would be to generalize RWR for multiplex-heterogeneous for $n$ multiplex networks and $n(n-1)/2$ bipartite linking them ($n \in \mathbb{N}$). Doing so, one could easily integrate many different types of nodes. The previous discussion about directed networks is in addition also valid for multiplex-heterogeneous network embedding.


By integrating different types of edges for multiplex network embedding or by integrating different types of both edges and nodes for multiplex-heterogeneous network embedding, MultiVERSE could have a wide variety of applications in diverse domains such as network biology and medicine, social science, computer science, neuroscience or physics. Our illustration of MultiVERSE embedding to study gene-disease associations could easily be applied to drug repositioning and drug discovery, for instance with a multiplex drug-drug network, a drug-target bipartite and a molecular multiplex. In this way, genes, diseases and drugs could be projected in the same vector space for further studies. In neuroscience, multiplex-heterogeneous network embedding could be applied to study the links between genes and neurons \cite{badhwar2015control}. In social science, multiplex networks are gaining interest to understand human behaviour \cite{smith2019using}. Multiplex-heterogeneous network embedding could give insights on epidemic spread \cite{johnson2015spillover, liu2018measurability}, socio-economic systems \cite{saracco2016detecting} or socio-ecological systems \cite{lenormand2018multiscale}.


\section*{Funding}

L.P-L. is the recipient of a Short Term Collaboration Grant for HPC 2019 from the Eurolab4HPC consortium. This project has  received  funding  from  the  Excellence  Initiative  of  Aix-Marseille   University- A*Midex, a French ‘Investissements d’Avenir’ program.

\section*{Conflict of Interest}

None declared.

\section*{Acknowledgements}

We thank Pr. Alfonso Valencia and the Computational Biology Life Sciences Group at Barcelona Supercomputing Center for the warm welcome and discussions and the use of the MareNostrum supercomputer.

\bibliography{example_paper}

\end{document}